\documentclass[letterpaper, 10 pt, conference]{ieeeconf}  % Comment this line out if you need a4paper
\IEEEoverridecommandlockouts                              % This command is only needed if 
\usepackage{graphics} % for pdf, bitmapped graphics files
\usepackage{epsfig} % for postscript graphics files
\usepackage{mathptmx} % assumes new font selection scheme installed
\usepackage{amsmath} % assumes amsmath package installed
\usepackage{amssymb}  % assumes amsmath package installed
\usepackage{lipsum}
\usepackage{multicol}
\usepackage{multirow}
\usepackage{booktabs}
\usepackage{stfloats}
\usepackage{float}
\usepackage{color,xcolor}
\usepackage{soul}
\usepackage{geometry}
\usepackage{hyperref}
\usepackage[export]{adjustbox} 

\usepackage{booktabs,tabularx}
\let\labelindent\relax
\usepackage{enumitem}

\usepackage{ulem}
\usepackage{amsmath,amssymb,booktabs,tabularx,siunitx,subcaption,graphicx,xcolor}

\begin{document}
\title{\LARGE \bf
ActionReasoning: Robot Action Reasoning in 3D Space with LLM for Robotic Brick Stacking
}

\author{Guangming Wang$^{*}$, Qizhen Ying$^{*}$, {Yixiong Jing}{\textdagger}, Olaf Wysocki, and Brian Sheil 
\thanks{This work was supported in part by Computer Vision for Digital Twins (CV4DT), Cambridge Centre for Smart Infrastructure and Construction (CSIC) and Laing O’Rourke Centre for Construction Engineering and Technology, Cambridge. (Corresponding Author: Yixiong Jing)}% <-this % stops a space
\thanks{* Equal contributions  {\textdagger} Co-corresponding authors} 
\thanks{G. Wang, Y. Jing, Olaf Wysocki, and B. Sheil are with the CV4DT, CSIC, Department of Engineering, University of Cambridge, Cambridge CB2 1PZ, U.K. (e-mail: gw462@cam.ac.uk, yj401@cam.ac.uk, okw24@cam.ac.uk, bbs24@cam.ac.uk)}%
\thanks{Q. Ying is with the Department of Engineering, University of Oxford, Oxford, U.K. (e-mail: qizhen.ying@exeter.ox.ac.uk). }
\thanks{Code will be available at 
\href{https://github.com/StephenYing/Action_Reasoning}{https://github.com/StephenYing/Action\_Reasoning}}
}

\maketitle

\begin{abstract}
Classical robotic systems typically rely on custom planners designed for constrained environments. While effective in restricted settings, these systems lack generalization capabilities, limiting the scalability of embodied AI and general‑purpose robots. Recent data‑driven Vision‑Language‑Action (VLA) approaches aim to learn policies from large‑scale simulation and real‑world data. However, the continuous action space of the physical world significantly exceeds the representational capacity of linguistic tokens, making it unclear if scaling data alone can yield general robotic intelligence. To address this gap, we propose ActionReasoning, an LLM-driven framework that performs explicit action reasoning to produce physics-consistent, prior-guided decisions for robotic manipulation. ActionReasoning leverages the physical priors and real-world knowledge already encoded in Large Language Models (LLMs) and structures them within a multi-agent architecture. We instantiate this framework on a tractable case study of brick stacking, where the environment states are assumed to be already accurately measured. The environmental states are then serialized and passed to a multi-agent LLM framework that generates physics-aware action plans. The experiments demonstrate that the proposed multi-agent LLM framework enables stable brick placement while shifting effort from low-level domain-specific coding to high-level tool invocation and prompting, highlighting its potential for broader generalization. This work introduces a promising approach to bridging perception and execution in robotic manipulation by integrating physical reasoning with LLMs.

\end{abstract}

\begin{figure}[t]
    \centering
\includegraphics[width=1\linewidth]{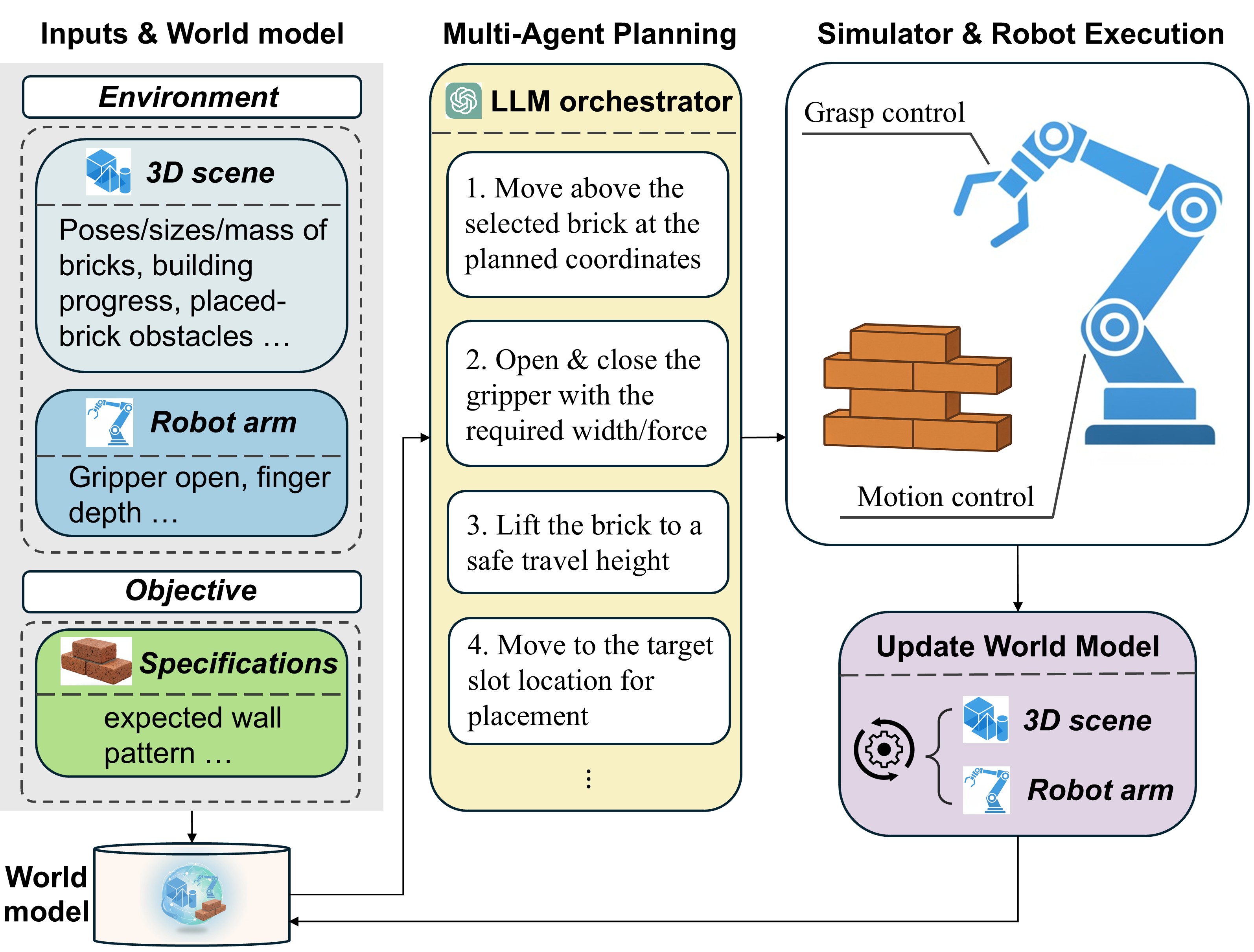}
    \caption{Three stages of the ActionReasoning pipeline. (1) Inputs \& World Model: The world model of bricklaying 
    is provided as input. (2) Multi-Agent Planning: Leveraging the world model input, an LLM orchestrator decomposes the task into specialized agents that generate actions and waypoints
    to plan the motion of a selected brick toward its target location. (3) Simulator \& Robot Execution: The robot (a Kuka simulator in this case) executes the planned actions from multiple agents to control grasp and motion. Observations from changes in the 3D scene and robot arm state are used to update the world model, enabling continual re-planning as the task progresses.}
    \label{fig:1}
\end{figure}

\section{INTRODUCTION}

Recent progress in humanoids and general manipulation policies has attracted significant attention in embodied intelligence. VLA models, such as RT‑2 \cite{brohan2023rt2} and OpenVLA \cite{kim2024openvla}, large open datasets like Open‑X‑Embodiment \cite{openxembodiment2023rtx}, and open policies such as Octo \cite{octo2024} demonstrate that scaling data and model capacity can improve reusability across tasks and hardware platforms. However, while LLMs in the text domain demonstrate emergent generalization through scaling, such “scale→emergence” capability has not yet been achieved with comparable robustness for universal robot control. Performance continues to degrade across diverse tasks, embodiments, and long-horizon reasoning, which indicates that merely scaling data and parameters alone is insufficient for robust generalization. Therefore, the proposition that “sheer scaling alone can yield general-purpose manipulation” needs to be treated with increasing caution in the robotics community, as there is a much larger solution space of robot actions compared to the language space. 

The lack of generalizability in robotics has motivated the increasing attention and research interest in world models. World models have been defined and studied in multiple ways. Some research focuses on predicting environmental dynamics and deducing future states in a latent space to support planning and control \cite{ha2018worldmodels, hafner2019planet}. Other recent works have introduced `generative foundation world models' \cite{hafner2020dreamer,bruce2024genie} that produce interactive environments from condition information, like one image or one sentence of the current environment. The term spans both an abstraction of universal physical laws and commonsense knowledge, as well as computable representations of scene geometry and dynamics in the current environment. Accordingly, this paper adopts an operational definition for robots: World Model = (A) universal physical knowledge and commonsense priors + (B) precise representations of the environment in which the robot operates. Here, (A) corresponds to a “general knowledge base,” while (B) aligns with Simultaneous Localization and Mapping (SLAM) \cite{cadena2016slam} and semantic geometry \cite{zhu2024sni}, where (B) enables the environment state to be queryable and constrained.

This paper focuses on how a world model can be used for robotics tasks, rather than how to learn one. As shown in Fig. \ref{fig:1}, we assume that the robot can obtain accurate 3D environmental states (e.g., 3D scene and robot arm states) through existing perception pipelines, and then combine these with the physical commonsense and task specifications (e.g., objective) already internalized in LLMs. This integration enables a form of 3D action reasoning: the LLM performs multi-step physical reasoning in the 3D environment with world model input, where the reasoning from the LLM orchestrator guides the robot arm in the simulator to finish tasks. To make the problem concrete and testable, we conduct systematic experiments and ablations on a representative task: brick stacking. This task involves typical physical constraints such as object contact, stability, and tolerance, while also requiring multi-step sequencing and precise pose coordination, making it a suitable testbed for 3D action reasoning.

The recent LLM-based robot system, ReKep \cite{huang2024rekep}, leveraged a generalizable perception models, Vision Language Models (VLMs), to identify actionable keypoints in the environment. However, 2D action keypoints lack understanding of complex 3D scenes, making them vulnerable to occlusion, foreshortening, and varied camera placements. Moreover, when LLMs are given images as inputs, these keypoints occupy a certain image region and are sometimes ambiguous or poorly localized. In addition, it still relies on dedicated off-the-shelf solvers. Our framework differs in that it (i) ingests explicit 3D scene states as LLM input, reducing reliance on 2D keypoint without 3D sensing, (ii) outputs phase‑conditioned action proposals and performs multi‑turn LLM calls when needed to refine plans, mimicking humans' “think‑while‑doing”.

Overall, this work achieves the following contributions:
\begin{itemize} 
\item We introduce ActionReasoning, i.e., a 3D action‑reasoning framework that enables robot decision‑making for brick stacking with high-level tool invocation and prompting, demonstrating generalization across configurations without per‑scene code.

\item We design a multi‑agent LLM orchestrator, where the specialized agents at different stages fuse human task specifications with world model input to reason in SE(3).

\item We provide simulation studies and ablations showing that 3D LLM reasoning improves robustness compared with the traditional control method with a similar simple programming, establishing a foundation for future LLM‑driven physical reasoning in robotics.

\end{itemize}

\section{RELATED WORK}
\subsection{Robotic Manipulation Learning}
\label{IIA}

Traditional manipulation robots rely on a structured pipeline in which the scene is perceived, a task-level representation is constructed, and detailed motions are programmed for execution within a given environment. Task‑and‑motion planning (TAMP) integrates symbolic task decomposition with continuous motion feasibility, but still assumes substantial domain modelling and engineering for each setting \cite{garrett2021tamp}. Even seemingly simple stacking tasks require extensive development on scripted sequencing, grasp poses, and constraint handling. 

Subsequently, Reinforcement Learning (RL) enabled training in environments without the need for task-specific programming for grasping or manipulation. By employing an action policy network and designing appropriate rewards, a robotic arm can explore the action space within simulations and achieve target tasks rewarded by the reward function. Since object positions (e.g., brick placements) can be freely configured in simulation, the success rate in simulated environments is often high. The subsequent research began exploring ways to accelerate this process, for example, by learning a corrective policy on top of a base controller \cite{johannink2019residual} or using base controllers as priors to guide exploration \cite{wang2022learning}. Such controllers bias the search toward regions of the solution space where conventional algorithms are more likely to succeed, thereby speeding up RL convergence. However, this still requires manually designing and coding the low-level controllers, with the ultimate aim of enabling RL to converge faster and eventually outperform the controllers themselves. Despite these advances, the sim-to-real gap remains a major challenge in this line of research \cite{wu2025rl}.

Although the recent Sim2Real2Sim pipeline proposed in \cite{wu2025rl} has made such sim‑to‑real transfers feasible, such trained models remain limited to the sampled real-world data distribution. Training a general model which is deployable in the real world still demands extensive real data collection and 3D modelling efforts. Similarly, VLA models \cite{brohan2023rt2,kim2024openvla,octo2024} push end‑to‑end learning from vision/language to actions, but also require large data. As a result, industry continues to rely heavily on large-scale data acquisition with demonstration datasets \cite{brohan2022rt1,walke2023bridgedata} and teleoperation datasets \cite{zhao2023aloha,aldaco2024aloha2}. These remain the mainstream solutions for improving the general capabilities of robots. However, a key challenge is that the robots’ action space is far larger than the vocabulary space of language. For LLMs, although the human vocabulary is vast, it is still finite. Once each word is vectorized, predicting the next word becomes a problem of discrete space prediction. In contrast, the action space of robots is enormously wide, raising the question of whether a scaling law exists and when it can arise for robots. 

\subsection{LLM/VLM Based Robotic Operation}

Thanks to access to big data, LLMs have achieved a form of general natural language question answering and have acquired substantial human-like understanding. Some might argue that LLMs have learned a mode of thinking similar to humans, while others insist that LLMs merely predict the distribution of the next token and essentially learn statistical patterns from data. The fact remains that LLMs can interpret and respond to general and intuitive questions. Moreover, they are increasingly capable of addressing domain-specific queries and drawing on broad knowledge of physics, mathematics, and other disciplines. The introduction of chain-of-thought reasoning has further enabled LLMs to follow human-preferred step-by-step reasoning processes in problem solving \cite{wei2022cot,yao2023treeofthought}. These advances have greatly enhanced the capabilities of LLMs, making them valuable for supporting robotic tasks such as task understanding, planning, and reasoning.

VLA models integrate visual understanding, language reasoning, and action generation to achieve action-level reasoning \cite{brohan2023rt2,kim2024openvla,octo2024}. However, end-to-end training requires massive datasets as we discussed in Section \ref{IIA}. Moreover, the training also introduces uncertainty at every stage, which spans from visual perception to linguistic reasoning and to final action generation. To address this, we argue for decoupling the visual component: assuming the robot already has sufficiently accurate perceptual information via computer vision algorithms, and the LLMs are asked to focus only on action reasoning. This approach significantly reduces data requirements while also leveraging the wealth of existing research in computer vision. For example, the ReKep series \cite{huang2024rekep, pan2025omnimanip} use vision-language models to identify keypoints for robotic manipulation, which significantly reduces the reliance on hand-coded programs and does not require training. Such approaches also constrain the robot’s exploration space.
The distinction of our work from ReKep \cite{huang2024rekep} is that while ReKep emphasizes 2D keypoint detection and corresponding actions, our method enables direct 3D spatial reasoning leveraging physical information. We therefore term it physical reasoning. This allows robots to exploit richer physical cues and directly operate in SE(3) to produce phase‑conditioned action proposals in 3D, avoiding limitations such as occlusions and the lack of precision inherent in 2D keypoint inputs.

In addition, for complex multi-step tasks, there are emerging multi-agent frameworks, such as CAMEL \cite{li2023camel} for role‑playing agents and AutoGen \cite{wu2023autogen} for multi‑agent conversations, that enable the coordination of specialized agents to debate, verify, or decompose tasks.
Embodied agents such as Voyager  \cite{wang2023voyager} demonstrate open‑ended skill acquisition with iterative prompting and a growing skill library. These patterns inform our design of specialized robotics agents, task decomposition, physics checking, and pose planning, while collaborating over a shared and updated 3D environment state. Our method employs multiple LLM calls, where each call is assigned specific roles to decompose complex problems, enabling end-to-end reasoning based on LLMs \cite{ghafarollahi2025automating}.

\section{METHODS}

\begin{figure*}
    \centering
    \includegraphics[width=\linewidth]{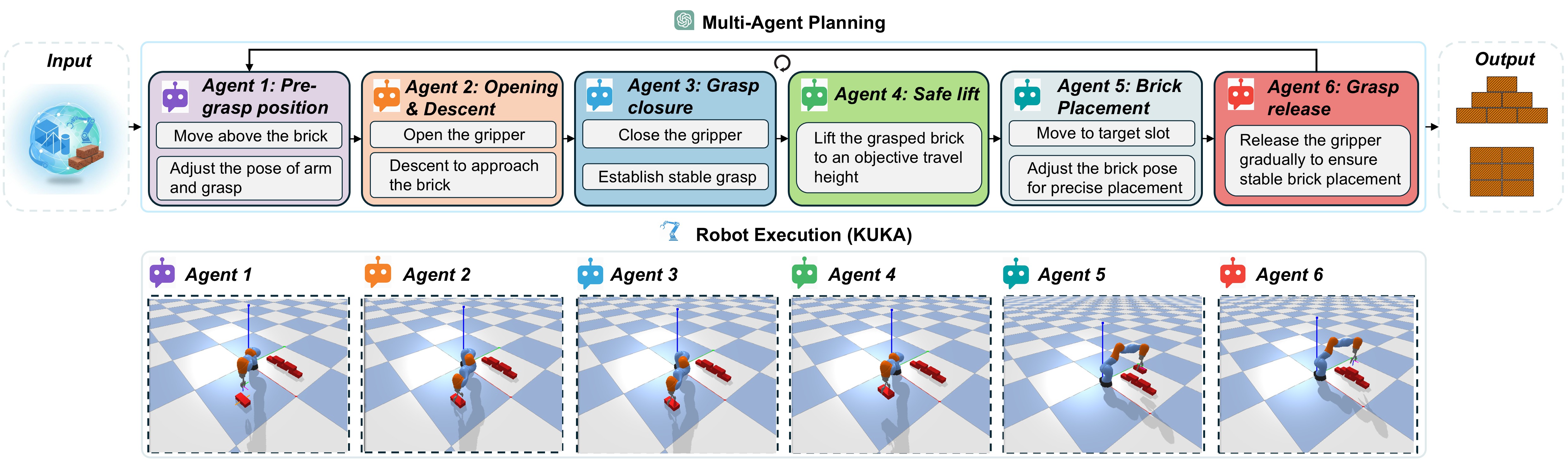}
    \caption{Illustration of the six agents ($Ag_1 - Ag_6$) in the present ActionReasoning framework: (1) Pre-grasp positioning to guide the arm to approach the brick; (2) Opening and descent to position the gripper above the brick surface; (3) Grasp closure to secure the brick; (4) Safe lift to raise the brick from the ground; (5) Brick placement to stably move and accurately align the brick at the target location; and (6) Grasp release to land the brick. 
    The corresponding execution of each agent is illustrated on the simulated KUKA robot arm at the bottom of this figure.}
    % \vspace{-3mm}
    \label{fig:pipeline}
\end{figure*}

We aim to bridge perception and execution for robotic manipulation via physical reasoning with LLMs. The input is the robot's structured understanding of the current environment state, denoted \( S_t \in \mathcal{S} \), including 3D scene geometry, brick poses, placed-brick obstacles, robot arm state, and task context, together with a goal specification \( G \in \mathcal{G} \). The output is an executable action for the manipulator, which is named as a waypoint, representing the pose of end-effector \( w_{t+1} \in \mathrm{SE}(3) \). Sometimes the output also includes the gripper opening and closing, but the following formula will omit this for the sake of simplicity. Low-level control, trajectory interpolation and high-rate control updates are handled by the robot controller embedded in the robot arm. Our objective is to use LLM with physics reasoning to close the gap between ``understanding'' and ``doing'' without a lot of low-level domain-specific coding.

\subsection{LLM based Multi-agent Framework for Robot Action Reasoning}\label{agent}
\subsubsection{Markovian waypoint loop}\label{Markovian}
Let \( A_t \in \mathcal{A} \) denote an action applied at time \( t \). The manipulator influences the environment through \( A_t \), producing a transition \( S_t \to S_{t+1} \). We run in waypoint mode: the method emits the next target pose \( w_{t+1} \in \mathrm{SE}(3) \). After each waypoint is executed, perception is refreshed, and the reasoner is called again. This yields a Markovian closed loop whose sampling granularity \( K_t \) adapts to task precision.
\begin{equation}
w_{t+1} = \pi_{\mathrm{AR}}(S_t, G) \in \mathrm{SE}(3),
\end{equation}
where LLM $\pi_{\mathrm{AR}}$ is used for 3D action reasoning to obtain the target pose $w_{t+1}$ from current
environment state $S_t$ and goal $G$.

Then, the sequence of low-level control commands $\mathbf{u}_{t,1:K_t}$ are interpolated over $K_t$ servo ticks to drive the robot from the current state $x_t$ toward the target waypoint $w_{t+1}$ as follows.
\begin{equation}
\mathbf{u}_{t,1:K_t} = \operatorname{Interp}\!\big(x_t \!\to\! w_{t+1};\, K_t\big),
\end{equation}
where \( \operatorname{Interp}(\cdot) \) is the internal trajectory generator. This is completed in the robot arm with default parameters and is not an academic contribution of this paper. Influenced by $\mathbf{u}_{t,1:K_t}$, the environment transits from $S_t$ to $S_{t+1}$ by the world transition \( \mathcal{T} \), which is simulated to reflect the physical laws in a real word.
\begin{equation}
S_{t+1} \sim \mathcal{T}\!\big(S_t, \mathbf{u}_{t,1:K_t}\big) 
\end{equation}
Then, the robot at pose $x_{t+1}$ perceives the updated environment to refresh to the structured environment state $\hat{S}_{t+1}$.
\begin{equation}
\hat{S}_{t+1} = \Phi\!\big(\operatorname{Perceive}(x_{t+1})\big) 
\end{equation}
To focus on our contribution in physical reasoning with LLMs, we adopt the simplification $\hat{S}_{t+1}={S}_{t+1}$.

Therefore, we obtain the Markov assumption that the next state depends only on the current state $S_t$ and the chosen waypoint $w_{t+1}$, rather than the full history $S_{0:t}$, which can be represented as follows.
\begin{equation}
p(S_{t+1}\!\mid S_{0:t}, G) = p(S_{t+1}\!\mid S_t, w_{t+1})
\end{equation}

\subsubsection{Feasible set and selection}
We cast action selection as a physics-guided inverse problem: given \( S_t \in \mathcal{S} \) and \( G \in \mathcal{G} \), infer an action (or waypoint) that makes goal-directed progress while satisfying safety and feasibility. Let \( F \) be a prior physical deduction operator encoding collision avoidance, reachability, contact stability, and tolerances.
\vspace{-3mm}
\begin{align}
\mathcal{A}_t^{\mathrm{feas}}
&= F(S_t, G) \nonumber \\
&=
\Big\{ a \in \mathcal{A}\;\Big|\;
\begin{aligned}[t]
&\underbrace{\mathcal{C}_{\mathrm{phys}}(S_t,a) \le 0}_{\text{physics/safety}},\; 
\underbrace{\mathcal{P}_{\mathrm{reach}}(S_t,a)=1}_{\text{reachability}},\\[2pt]
&\underbrace{\Delta(S_t,a;G) \le \varepsilon}_{\text{goal progress}}
\end{aligned}
\Big\}, \label{eq:feasible} \\
a_t^\star &= \arg\min_{a \in \mathcal{A}_t^{\mathrm{feas}}} J(a; S_t, G),
\label{eq:argmin}
\end{align}
% \vspace{-1mm}
where \( \mathcal{C}_{\mathrm{phys}} \) captures collision/contact limits, \( \mathcal{P}_{\mathrm{reach}} \) is a reachability predicate (including kinematic and joint limits), \( \Delta \) measures residual error relative to \( G \), \( \varepsilon \) is a tolerance, and \( J \) trades off path length, clearance, and alignment quality. In practice, the LLM proposes candidates consistent with priors, and a verifier prunes them using \( F \).

\subsubsection{Agent pipeline with gating}
Long-horizon brick stacking is handled by a sequential gated multi-agent pipeline. Each agent \( Ag_i \) consumes the current state and upstream messages, outputs a message \( m_i \) (e.g., a pose proposal, constraints, or a waypoint), and returns an acceptance flag \( \sigma_i \in \{0,1\} \). The next agent executes only if the previous one accepts.

As shown in Fig.~2, we instantiate six specialized agents \( Ag_1 \)–\( Ag_6 \) that transform high-level intent into physically consistent motion targets. We can formulate them simply as: 
\begin{align}
(m_1,\sigma_1) &= Ag_1(S_t, G), \label{eq:ag1} \\
(m_2,\sigma_2) &= Ag_2(S_t, G, m_{1}), \label{eq:ag2} \\
(m_3,\sigma_3) &= Ag_3(S_t, G, m_{1:2}), \label{eq:ag3} \\
(m_4,\sigma_4) &= Ag_4(S_t, G, m_{1:3}), \label{eq:ag4} \\
(m_5,\sigma_5) &= Ag_5(S_t, G, m_{1:4}), \label{eq:ag5} \\
(m_6,\sigma_6) &= Ag_6(S_t, G, m_{1:5}), \label{eq:ag6} \\[3pt]
A_t &= 
\begin{aligned}[t]
\big(Ag_6 \circ Ag_5 \circ Ag_4 \circ Ag_3 \circ Ag_2 \circ Ag_1\big)(S_t, G)
\end{aligned} \notag \\
&\qquad \text{iff} \quad \sigma_i=1~\forall i\in\{1,\dots,6\}.
\label{eq:compose}
\end{align}
The gating rule can be written as
\begin{equation}
Ag_{i+1}\ \text{is executed iff }\ \sigma_i
= \mathbf{1}\!\left[\operatorname{Checks}_i\big(S_t, G, m_{1:i}\big)\right] = 1.
\label{eq:gating}
\end{equation}
Specifically, for each agent, the design is as follows. Below, \( \hat{n} \) denotes a surface normal, \( f_n \) is a measured normal force, \( \mu \) is an estimated friction coefficient, and \( \mathbf{e}_{xy} \) and \( e_{\theta} \) denote the planar translation and yaw alignment errors, respectively.

\paragraph{\( Ag_1 \) Pre-grasp positioning}
The agent moves the robot end-effector above the target brick and aligns the gripper approach vector with the brick’s grasp axis. The agent proposes an approach pose \( p_{\mathrm{app}} \in \mathrm{SE}(3) \) with clearance \( c \ge c_{\min} \):
\begin{equation}
\sigma_1 = \mathbf{1}\!\left[
\operatorname{CollisionFree}\!\big(\operatorname{Path}(x_t \!\to\! p_{\mathrm{app}})\big)
\land c \ge c_{\min}
\right].
\label{eq:ag1check}
\end{equation}

\paragraph{\( Ag_2 \) Descent \& opening}
The agent opens the gripper to width \( w \ge w_{\mathrm{brick}} + \delta \) and descends along \( -\hat{z} \) into a graspable pose \( p_{\downarrow} \):
\begin{equation}
\sigma_2 = \mathbf{1}\!\left[
w \ge w_{\mathrm{brick}} + \delta
\land \operatorname{NoContactSidewalls}(p_{\downarrow})
\right].
\label{eq:ag2check}
\end{equation}

\paragraph{\( Ag_3 \) Grasp closure}
The agent closes the gripper and verifies a stable contact using a normal-force threshold and pose tolerance:
\begin{equation}
\sigma_3 = \mathbf{1}\!\left[
f_n \ge f_{\min}
\land \operatorname{PoseError}(p_{\downarrow},p_{\mathrm{grasp}})\le \varepsilon_g
\right].
\label{eq:ag3check}
\end{equation}

\paragraph{\( Ag_4 \) Safe lift}
The agent lifts vertically to height \( h_{\mathrm{safe}} \) and evaluates slip risk. If risky, it returns to \( Ag_1 \) for re-grasp:
\begin{equation}
\sigma_4 = \mathbf{1}\!\left[
\|v_{\mathrm{brick}}\| \le v_{\mathrm{th}}
\land f_t \le \mu f_n
\right],
\qquad
\sigma_4 = 0 \Rightarrow \text{goto } Ag_1.
\label{eq:ag4check}
\end{equation}

\paragraph{\( Ag_5 \) Brick placement}
The agent moves to the designated slot pose \( p_{\mathrm{slot}} \in \mathrm{SE}(3) \), descends until contact, and checks alignment. If misaligned but in contact, raise by \( \Delta h \) and retry \( Ag_5 \):
\begin{align}
\sigma_5 &= \mathbf{1}\!\left[
d_{\perp} \le \varepsilon_{\perp}
\land \|\mathbf{e}_{xy}\| \le \varepsilon_{xy}
\land e_{\theta} \le \varepsilon_{\theta}
\right],
\label{eq:ag5check1} \\
\sigma_5 &= 0 \Rightarrow
\text{raise by } \Delta h \ \text{and repeat } Ag_5.
\label{eq:ag5check2}
\end{align}

\paragraph{\( Ag_6 \) Return-to-ready}
The agent retracts to a collision-free ready pose \( p_{\mathrm{ready}} \in \mathrm{SE}(3) \) for the next brick:
\begin{equation}
\sigma_6 = \mathbf{1}\!\left[
\operatorname{CollisionFree}\!\big(\operatorname{Path}(p_{\mathrm{slot}}\!\to\! p_{\mathrm{ready}})\big)
\right].
\label{eq:ag6check}
\end{equation}

When all \( \sigma_i = 1 \), \( Ag_6 \) emits the next waypoint \( w_{t+1} = p_{\mathrm{ready}} \), which is passed to the controller. The loop in Section \ref{Markovian} then updates perception \( S_{t+1} \) and repeats for the next brick.

This organization treats 3D action reasoning as a physics-guided inverse problem executed by a sequential gated LLM multi-agent pipeline. Waypoint control provides a safe and Markovian interface to the embedded controller. The operator \( F \) enforces physical constraints, while the LLM supplies structured proposals that exploit high-level priors. The proposed pipeline achieves using high-level tool invocation and prompting yet physically consistent manipulation for brick stacking.

\begin{figure*}[t]
    \centering
\includegraphics[width=1.0\linewidth]       {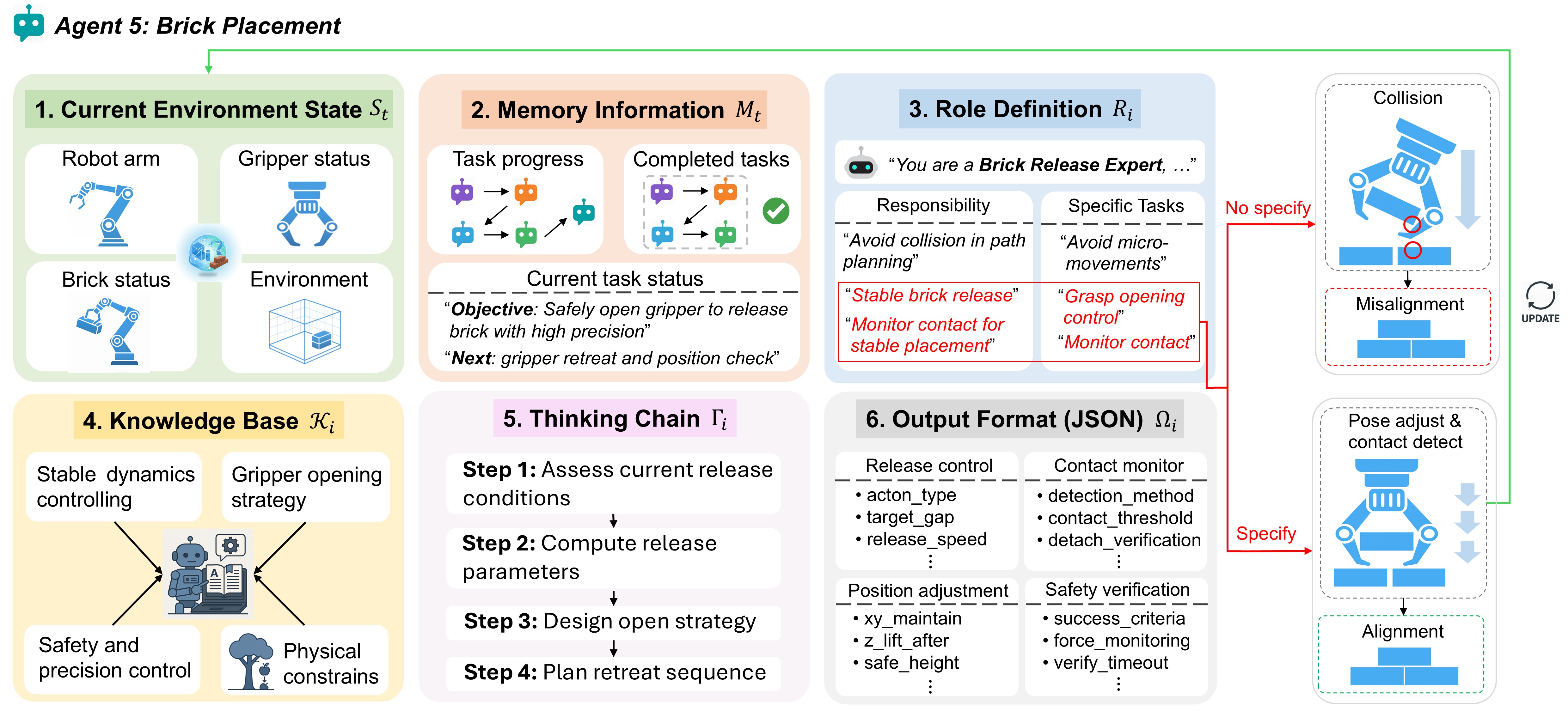}
    \caption{Detailed architecture of Agent 5 (Brick Placement) with six prompt-driven components. (1) Current environment state: provides the latest world model; (2) Memory information: explains task progress of previous agents and the current task status; (3) Role definition: specifies the agent’s function and responsibilities, including collision-avoidance tasks for stable brick placement. A comparison of the brick laying between specifying and not specifying is visualized aside; (4) Knowledge base: describes domain knowledge such as dynamics, gripper strategies, and safety constraints; (5) Thinking chain: outlines stepwise reasoning for placement and retreat; and (6) Output format: structured JSON commands for execution in the KUKA simulator.}
    \label{fig:somebigs}
\end{figure*} 

\subsection{Agent Construction via Structured Prompting}
\label{sec:agent-construction}

We instantiate one agent through a structured prompt that binds environment information, memory, role, tool knowledge, chain-of-thought guidance, and an explicit output schema. Fig.~3 gives one example for \( Ag_5 \) Brick placement.  Formally, agent \( Ag_i \) at time \( t \) receives:
\begin{equation}
\mathsf{Prompt}_i(t)
\;\triangleq\;
\big\langle
S_t,\; M_t,\; R_i,\; \mathcal{K}_i,\; \Gamma_i,\; \Omega_i
\big\rangle,
\end{equation}
where \( S_t \in \mathcal{S} \) is the current structured scene state, \( M_t \) is task memory, \( R_i \) is the role description, \( \mathcal{K}_i \) is a set of callable knowledge/tools, \( \Gamma_i \) is the chain-of-thought scaffold, and \( \Omega_i \) defines the output schema. Specifically, each module of the structured prompt can be described as follows.
\begin{itemize}
  \item \textbf{Current environment state} \( S_t \): robot and surroundings, including robot status, object poses/geometry, occupancy/free space, surface normals, and tolerance parameters.
  \item \textbf{Memory information} \( M_t \): current task status, task progress and completed tasks, e.g., completed bricks, current brick index, current step for current brick, retry counters, step completion flags.
  \item \textbf{Role definition} \( R_i \): concise instruction of the agent’s responsibility and task specifications, including what to accomplish and how it interfaces with other agents.
  \item \textbf{Knowledge base} \( \mathcal{K}_i \): basic knowledge in robotics, such as stable dynamics control method, safety and precision control strategies,  gripper opening strategies, and physical constraint solving. These are memoried in callable functions with typed I/O and usage (e.g., collision and contact checks, alignment estimators, safety margins).
  \item \textbf{Thinking Chain} \( \Gamma_i = (\gamma_i^{(1)},\ldots,\gamma_i^{(L_i)}) \): a stepwise reasoning script describing how to think and decide.
  \item \textbf{Output Format} \( \Omega_i \): JSON-like fields for waypoints in \( \mathrm{SE}(3) \) or code snippets to call tools.
\end{itemize}

The agent returns a tuple:
\begin{equation}
(m_i,\; y_i,\; \sigma_i,\; M_{t+1})
\;=\;
Ag_i\!\big(\mathsf{Prompt}_i(t)\big),
\end{equation}
where \( m_i \) is the agent’s message, including intermediate rationale or parameters, \( y_i \in \Omega_i \) is the actionable output, \( \sigma_i \in \{0,1\} \) is an acceptance flag or gate, and \( M_{t+1} \) is the updated memory. In our setting,
\[
y_i \in \Omega_i \;\subseteq\; \big(\mathrm{SE}(3)\ \cup\ \mathcal{C}\big),
\]
i.e., either a waypoint \( w \in \mathrm{SE}(3) \) for the manipulator or a small code snippet \( c \in \mathcal{C} \) to invoke a tool. Typical tools in \( \mathcal{K}_i \) include
collision check, reachability, normal/contact force estimate, and planar translation and yaw errors.
The gate \( \sigma_i \) is computed against explicit checks as Eq.~\eqref{eq:gating}.

\newcommand{\imgph}[2]{%
  \fbox{\begin{minipage}[c][#2][c]{#1}\centering\footnotesize Placeholder\end{minipage}}%
}
\sisetup{round-mode=places,round-precision=1}

\begin{table*}[t]
  \centering
  \caption{Comparison with classical baseline on two stacking patterns. Report rotation error (°), center offset (cm), and 3D box IoU (\%). For rotation error and center offset, the lower the better. For 3D box IoU, the higher the better.}
  \setlength{\tabcolsep}{4pt}
  \label{tab:baseline-compare}
  \newcolumntype{Y}{>{\centering\arraybackslash}X}
  \begin{tabularx}{\textwidth}{@{}l Y Y Y Y Y Y Y Y Y@{}}
    \toprule
    \multirow{2}{*}{\textbf{Method}} 
      & \multicolumn{3}{c}{\textbf{Pyramid-like stacking}} 
      & \multicolumn{3}{c}{\textbf{Grid-like stacking}}
      & \multicolumn{3}{c}{\textbf{Average}} \\
    \cmidrule(lr){2-4} \cmidrule(lr){5-7} \cmidrule(lr){8-10}
      & \textbf{Rot. Err $\scriptsize\downarrow$} & \textbf{Ctr. Off. $\scriptsize\downarrow$} & \textbf{IoU $\scriptsize\uparrow$}
      & \textbf{Rot. Err $\scriptsize\downarrow$} & \textbf{Ctr. Off. $\scriptsize\downarrow$} & \textbf{IoU $\uparrow$}
      & \textbf{Rot. Err $\scriptsize\downarrow$} & \textbf{Ctr. Off. $\scriptsize\downarrow$} & \textbf{IoU $\scriptsize\uparrow$} \\
    \midrule
    Classical Controller (baseline)  
      &1.103 & 4.318 & 38.51 
      & 0.939 & 4.379 & 37.72 
      & 1.004 & 4.314 & 38.38 \\
    ActionReasoning (Ours)  
       & \bf0.583 &  \bf0.561 & \bf89.03
      & \bf0.822 & \bf0.712 & \bf87.02
      & \bf0.703 & \bf0.637 & \bf88.03 \\
    \bottomrule
  \end{tabularx}
\end{table*}

\begin{figure*}[t]
  \centering

  % ---------- Left: Pyramid-like ----------
  \begin{minipage}{0.48\textwidth}\centering
    \textbf{Pyramid-like stacking}\\[0.5em]

    \textit{Classical Controller (Baseline)}\\[0.35em]
    \includegraphics[width=0.31\linewidth,height=0.31\linewidth]{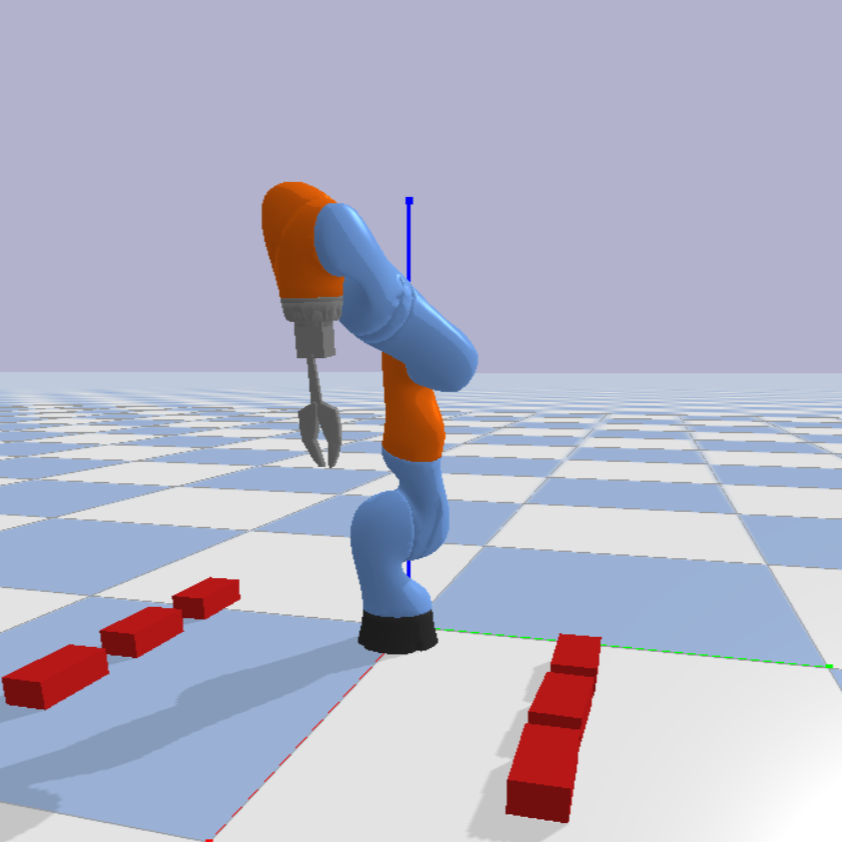}\hfill
    \includegraphics[width=0.31\linewidth,height=0.31\linewidth]{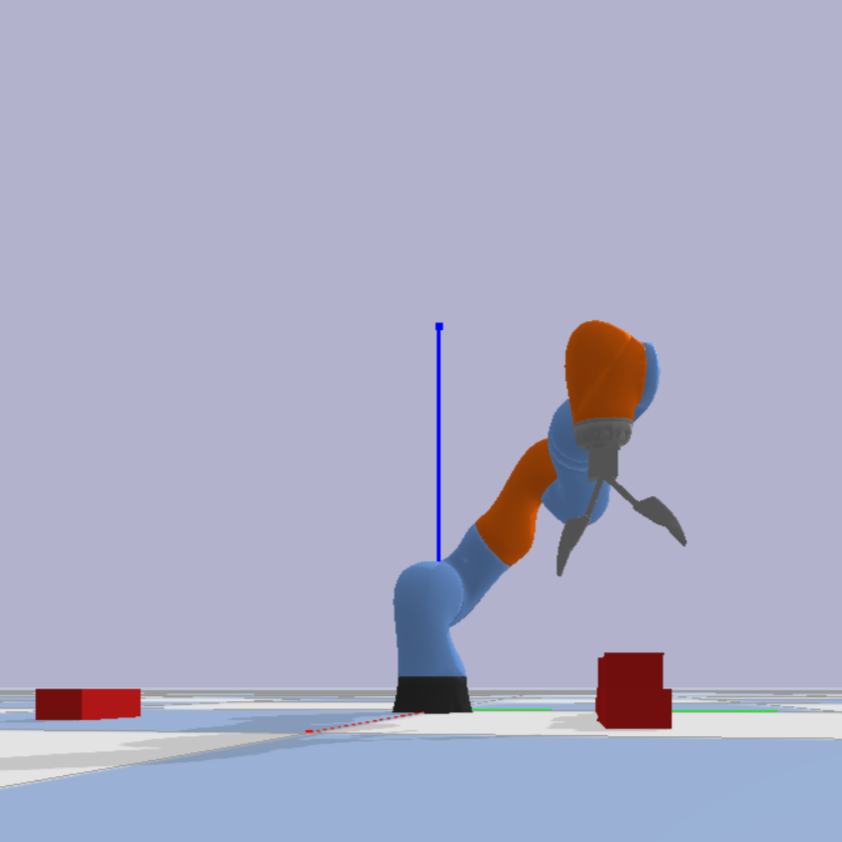}\hfill
    \includegraphics[width=0.31\linewidth,height=0.31\linewidth]{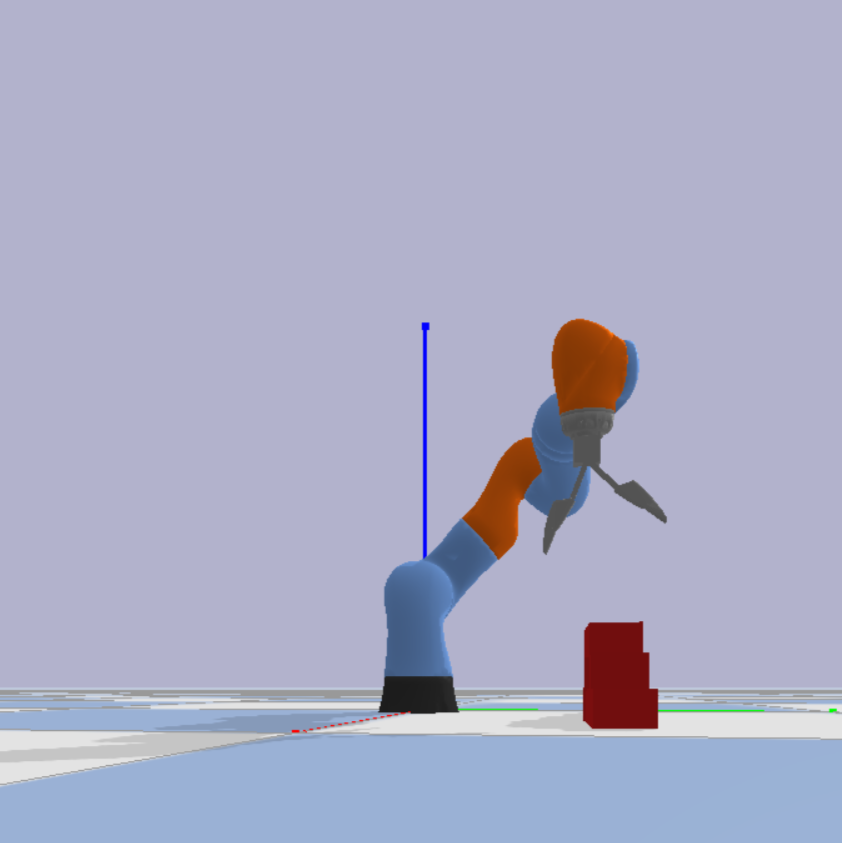}\\[0.45em]

    \textit{ActionReasoning (Ours)}\\[0.35em]
    \includegraphics[width=0.31\linewidth,height=0.31\linewidth]{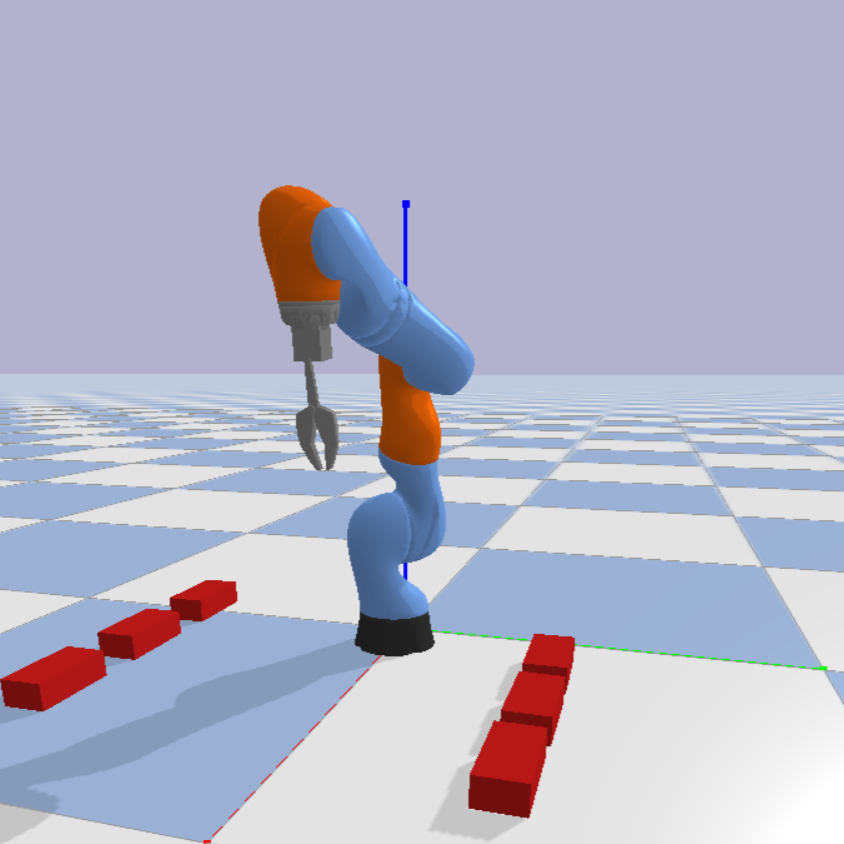}\hfill
    \includegraphics[width=0.31\linewidth,height=0.31\linewidth]{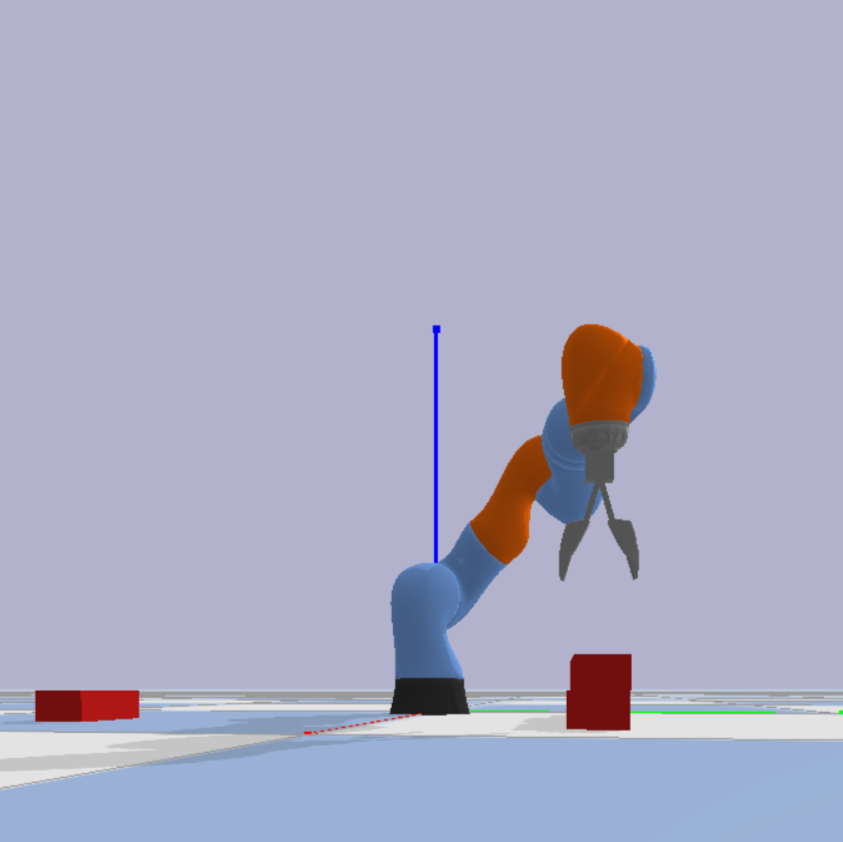}\hfill
    \includegraphics[width=0.31\linewidth,height=0.31\linewidth]{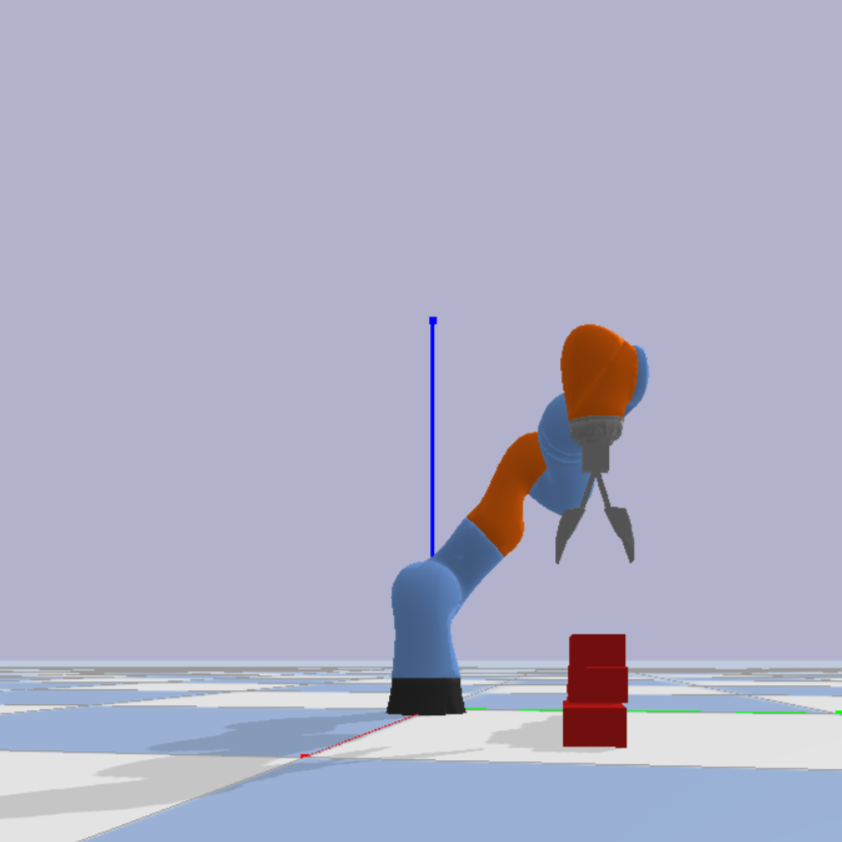}
  \end{minipage}
  \hfill
  % ---------- Right: Grid-like ----------
  \begin{minipage}{0.48\textwidth}\centering
    \textbf{Grid-like stacking}\\[0.5em]

    \textit{Classical Controller (Baseline)}\\[0.35em]
    \includegraphics[width=0.31\linewidth,height=0.31\linewidth]{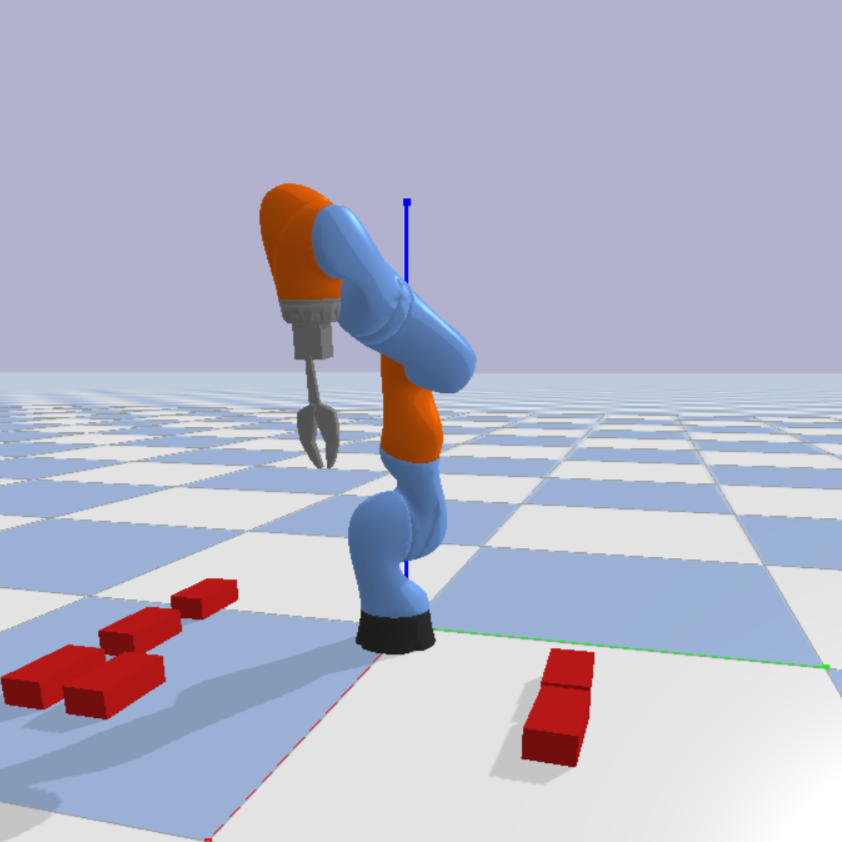}\hfill
    \includegraphics[width=0.31\linewidth,height=0.31\linewidth]{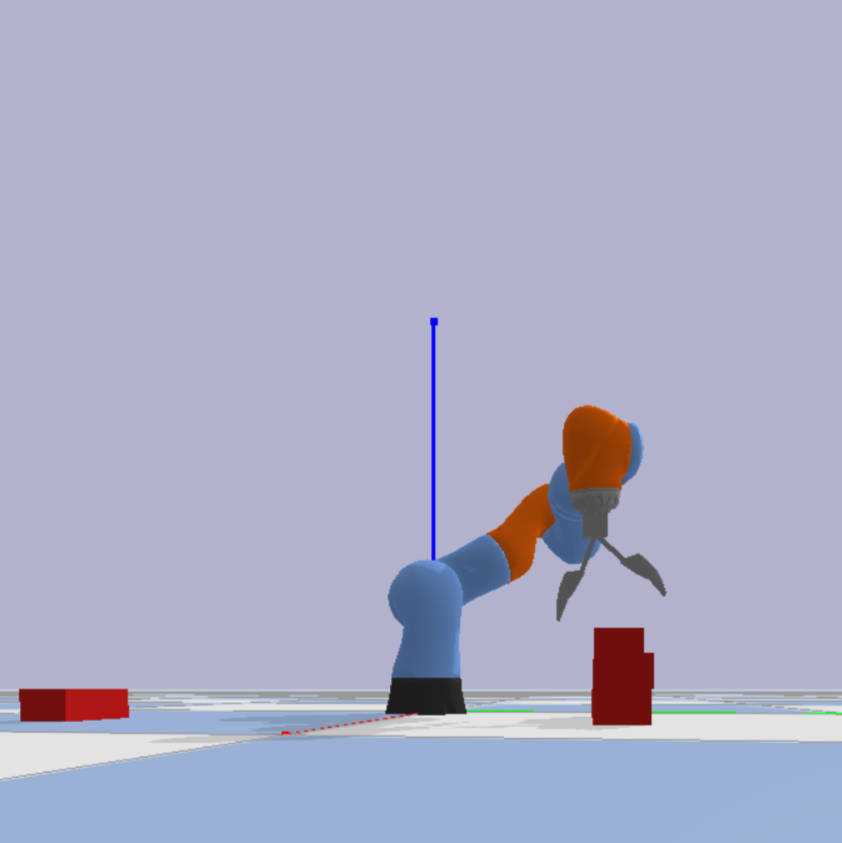}\hfill
    \includegraphics[width=0.31\linewidth,height=0.31\linewidth]{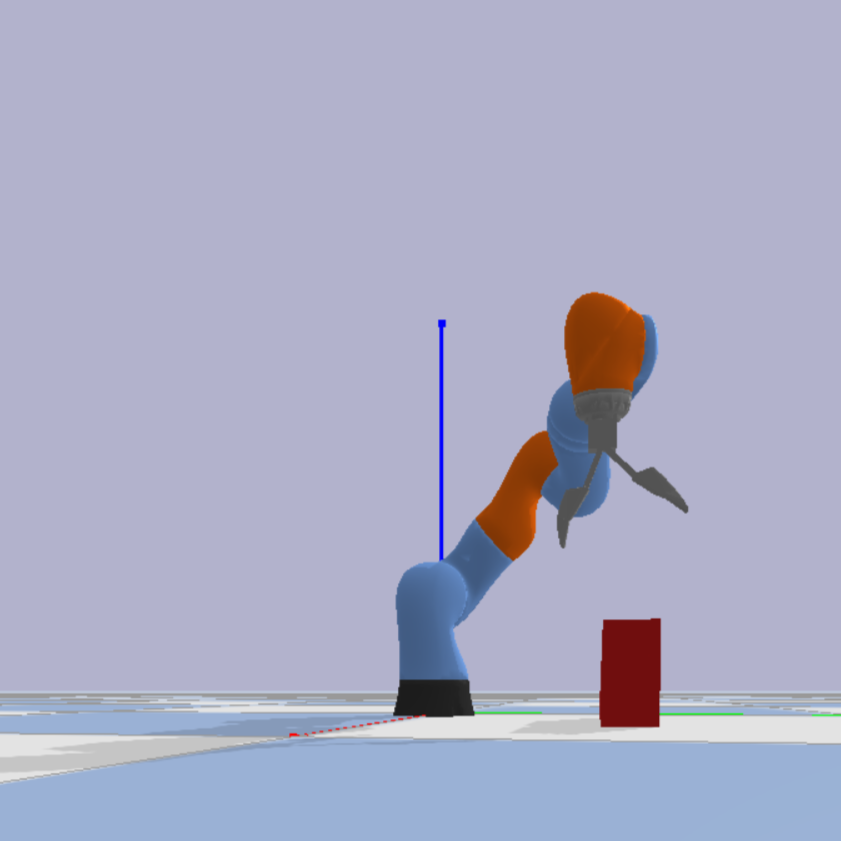}\\[0.45em]

    \textit{ActionReasoning (Ours)}\\[0.35em]
    \includegraphics[width=0.31\linewidth,height=0.31\linewidth]{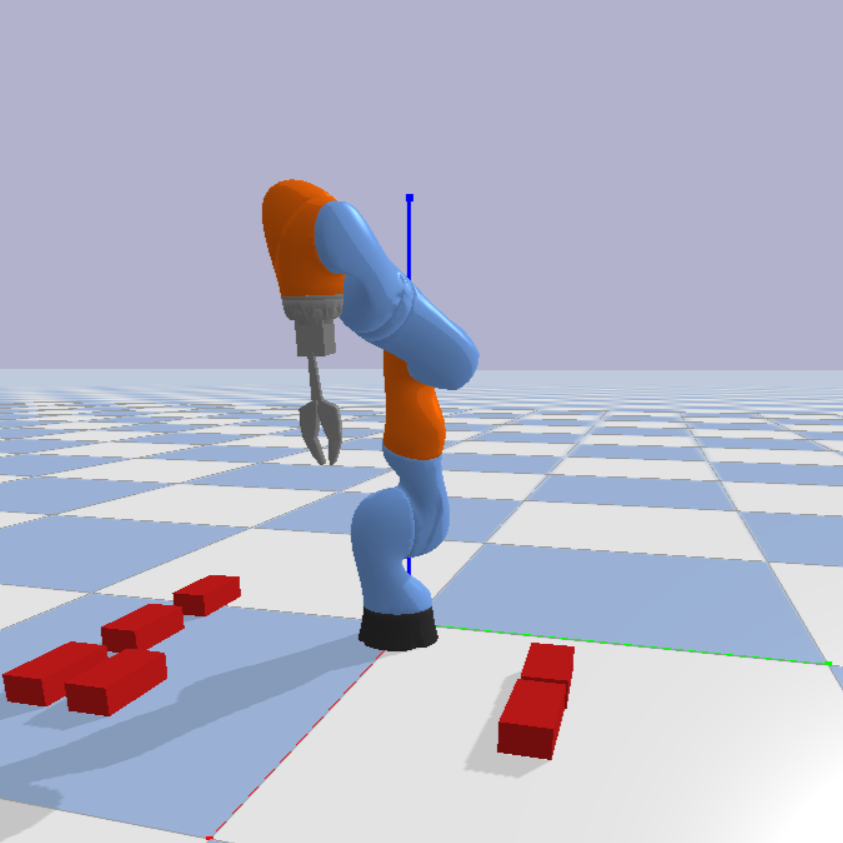}\hfill
    \includegraphics[width=0.31\linewidth,height=0.31\linewidth]{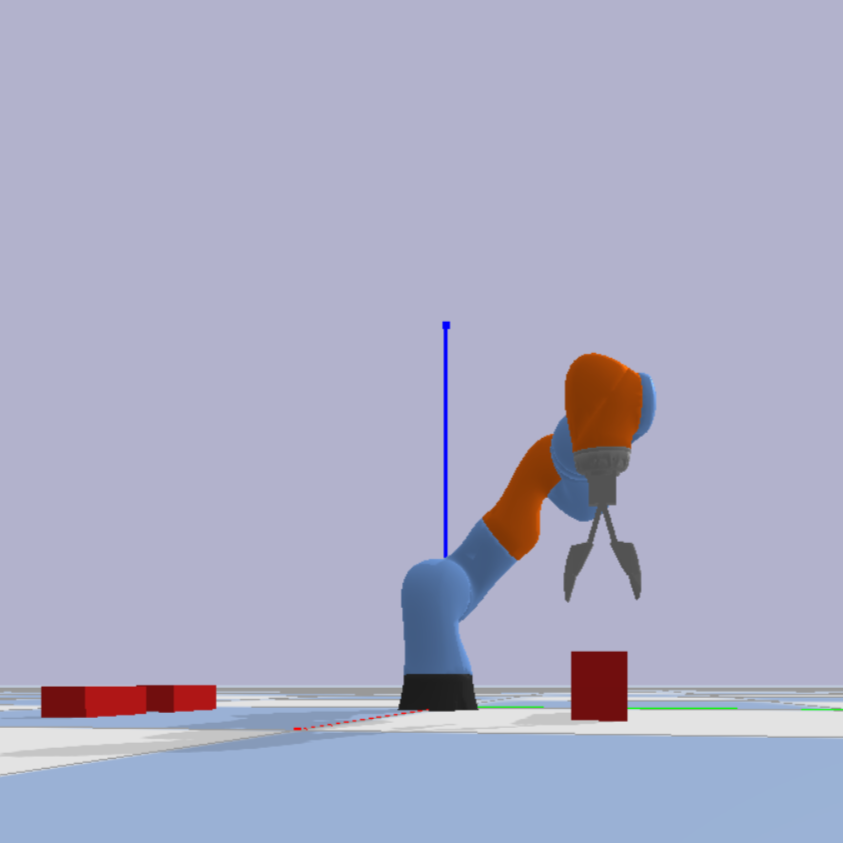}\hfill
    \includegraphics[width=0.31\linewidth,height=0.31\linewidth]{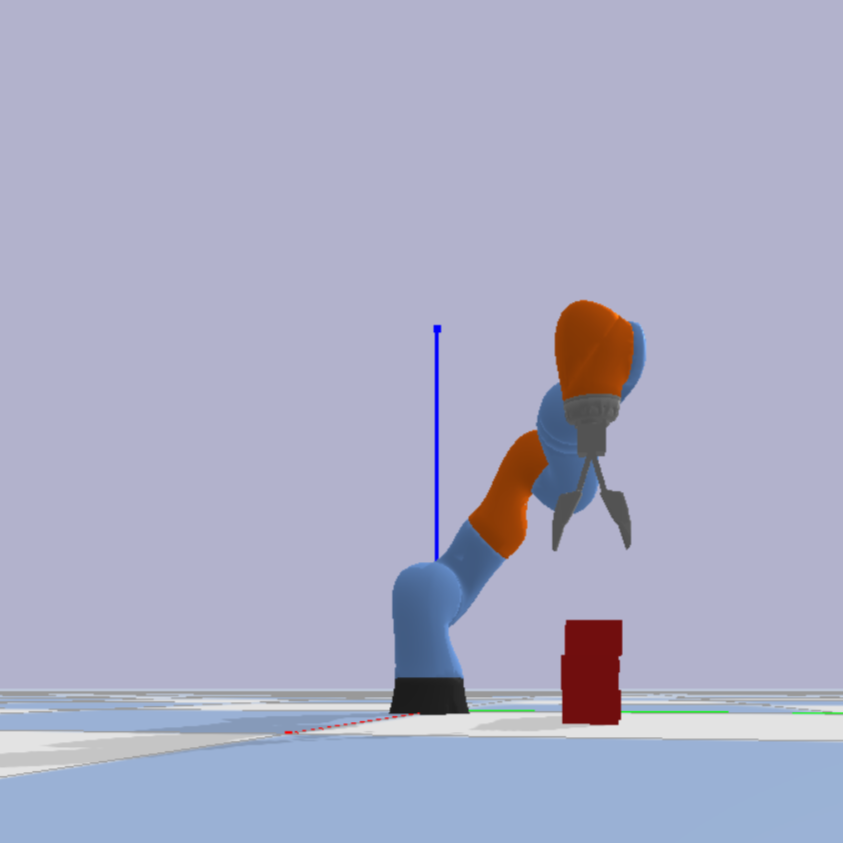}
  \end{minipage}

  \caption{Qualitative comparison across two stacking patterns. 
  Each block shows the baseline (top row) and our method (bottom row). 
  In each block of stacking pattern, columns show some stages of the placement cycle according to the timestamp from left to right.
  Our method achieves noticeably better brick alignment, as evidenced by the more neatly stacked bricks.}
  \label{fig:baseline-viz}
\end{figure*}

\section{Experiments}
\label{sec:experiments}

\subsection{Task and Setup}
\label{subsec:task-setup}
As illustrated in Fig.~\ref{fig:1}, the input to our system is a structured environment representation comprising: (i) a \(3\)D scene state with brick poses/sizes/masses, current building progress, and placed-brick obstacles; (ii) the robot-arm state, including gripper openness and finger depth; and (iii) a target wall pattern \( G \) specifying the expected arrangement. Formally, at time \( t \) we observe \( S_t \in \mathcal{S} \) and a fixed goal \( G \in \mathcal{G} \). The proposed multi-agent planner in Section \ref{agent} produces executable waypoints \( w_{t+1} \in \mathrm{SE}(3) \), and the embedded controller performs trajectory interpolation. After each action, the simulator updates the scene state to \( S_{t+1} \), and planning proceeds to the next step.

To evaluate generality, we randomize the initial brick poses within the workspace \( \mathcal{W} \subset \mathbb{R}^3 \).
We stack a total of \(6\) bricks to realize two types of patterns \( \mathcal{G} \) as shown in the output part of Fig. \ref{fig:pipeline}: (1) a pyramid-like stacking, where each upper layer has fewer bricks centered on those below to form a triangular shape, with a gap of 0.05 m between adjacent bricks, (2) a grid-like stacking, where bricks are aligned in a regular rectangular grid with vertical and horizontal joints aligned, with a gap of 0.02 m between adjacent bricks. The operation is performed entirely in the physics-based simulator Pybullet \cite{coumans2016pybullet}. Unless otherwise stated, contact/friction parameters and sensor noise follow the simulator defaults.

\subsection{Evaluation Protocol and Metrics}
\label{sec:metrics}

We evaluate the mentioned two target wall patterns in the Section \ref{subsec:task-setup}, each stacking \( B=6 \) bricks per trial. For each pattern, we conduct \( T_p=10 \) trials under randomized initial brick poses, yielding a total of \( \sum_{p} T_p = 20 \) trials. We report 
% the average success rate and 
the alignment quality metrics over all 20 experiments.

\subsubsection{Pose accuracy}
For each successful trial \( r \) and brick \( b \), the placed center and ground-truth center are denoted  \( \mathbf{c}_{r,b} \) and  \( \mathbf{c}^\star_{r,b} \), respectively. The center offset error is
\begin{equation}
e^{\mathrm{ctr}}_{r,b} \;=\; \big\| \mathbf{c}_{r,b} - \mathbf{c}^\star_{r,b} \big\|_2.
\end{equation}
The rotation error (in degrees) uses the geodesic distance on \( \mathrm{SO}(3) \) with rotation matrices \( R_{r,b}, R^\star_{r,b} \):
\begin{equation}
e^{\mathrm{rot}}_{r,b}
\;=\;
\frac{180}{\pi}\;
\arccos\!\left(\frac{\mathrm{trace}\!\big(R^{\star\top}_{r,b} R_{r,b}\big)-1}{2}\right).
\end{equation}
\subsubsection{3D IoU} Let \( B_{r,b} \) and \( B^\star_{r,b} \) be the oriented 3D bounding boxes of the placed and ground-truth bricks, respectively. The 3D intersection-over-union (IoU) can be computed as:
\begin{equation}
\mathrm{IoU}_{r,b}
\;=\;
\frac{\mathrm{Vol}\!\big(B_{r,b} \cap B^\star_{r,b}\big)}
     {\mathrm{Vol}\!\big(B_{r,b} \cup B^\star_{r,b}\big)}.
\end{equation}
Per-success trial averages are calculated over \( B=6 \) bricks:
\begin{equation}
\bar{e}^{\mathrm{ctr}}_{r}=\frac{1}{B}\sum_{b=1}^{B} e^{\mathrm{ctr}}_{r,b},\quad
\bar{e}^{\mathrm{rot}}_{r}=\frac{1}{B}\sum_{b=1}^{B} e^{\mathrm{rot}}_{r,b},\quad
\overline{\mathrm{IoU}}_{r}=\frac{1}{B}\sum_{b=1}^{B} \mathrm{IoU}_{r,b}.
\end{equation}
Global means over all successful trials \( \mathcal{R} \) are then:
{{\small
\begin{equation}
\mathrm{Err}^\mathrm{ctr}=\frac{1}{|\mathcal{R}|}\sum_{r\in\mathcal{R}}\bar{e}^{\mathrm{ctr}}_{r},\quad
\mathrm{Err}^\mathrm{rot}=\frac{1}{|\mathcal{R}|}\sum_{r\in\mathcal{R}}\bar{e}^{\mathrm{rot}}_{r},\quad
\mathrm{IoU}=\frac{1}{|\mathcal{R}|}\sum_{r\in\mathcal{R}}\overline{\mathrm{IoU}}_{r}.
\label{eq:agg_metrics}
\end{equation}}

\begin{figure}[t]
  \centering
  \begin{subfigure}{0.5\textwidth}\centering
    \textbf{Single-Agent (Ablation study)}\\[0.3em]
\includegraphics[width=0.31\linewidth]{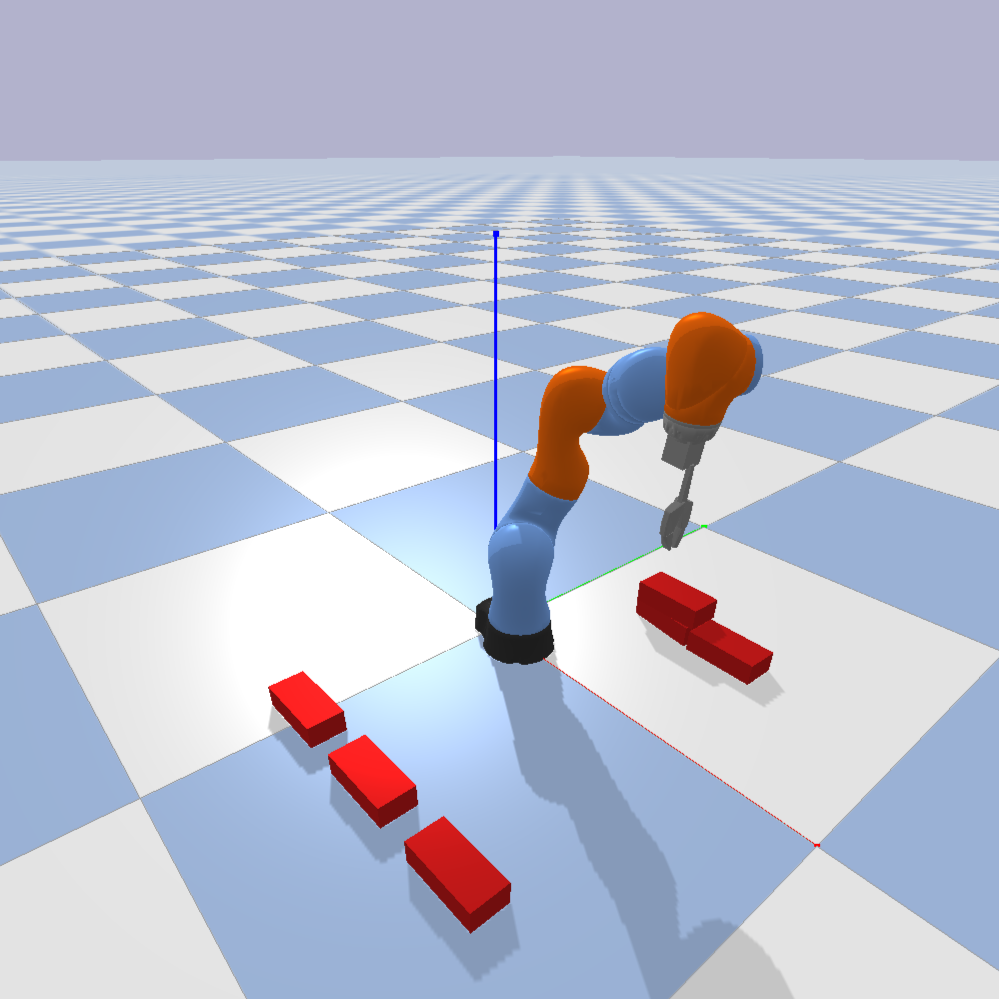}\hfill \includegraphics[width=0.31\linewidth]{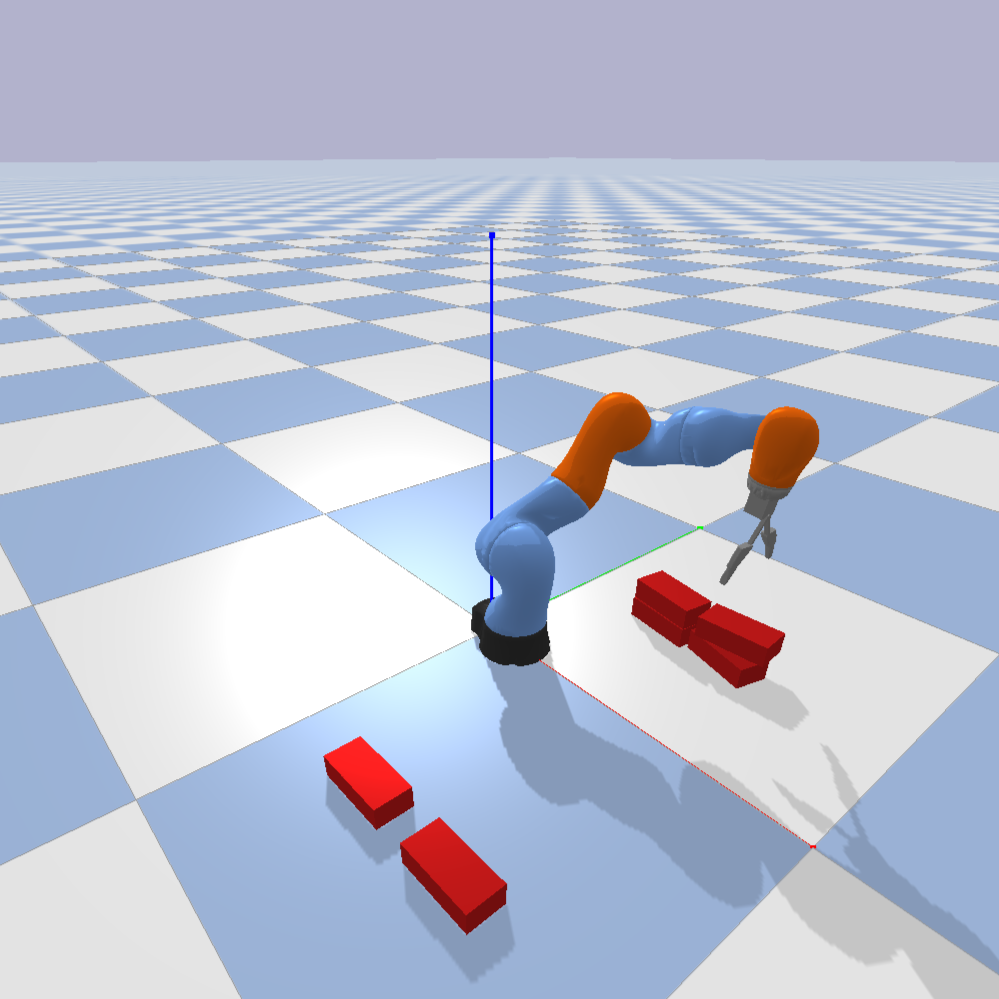}\hfill
\includegraphics[width=0.31\linewidth]{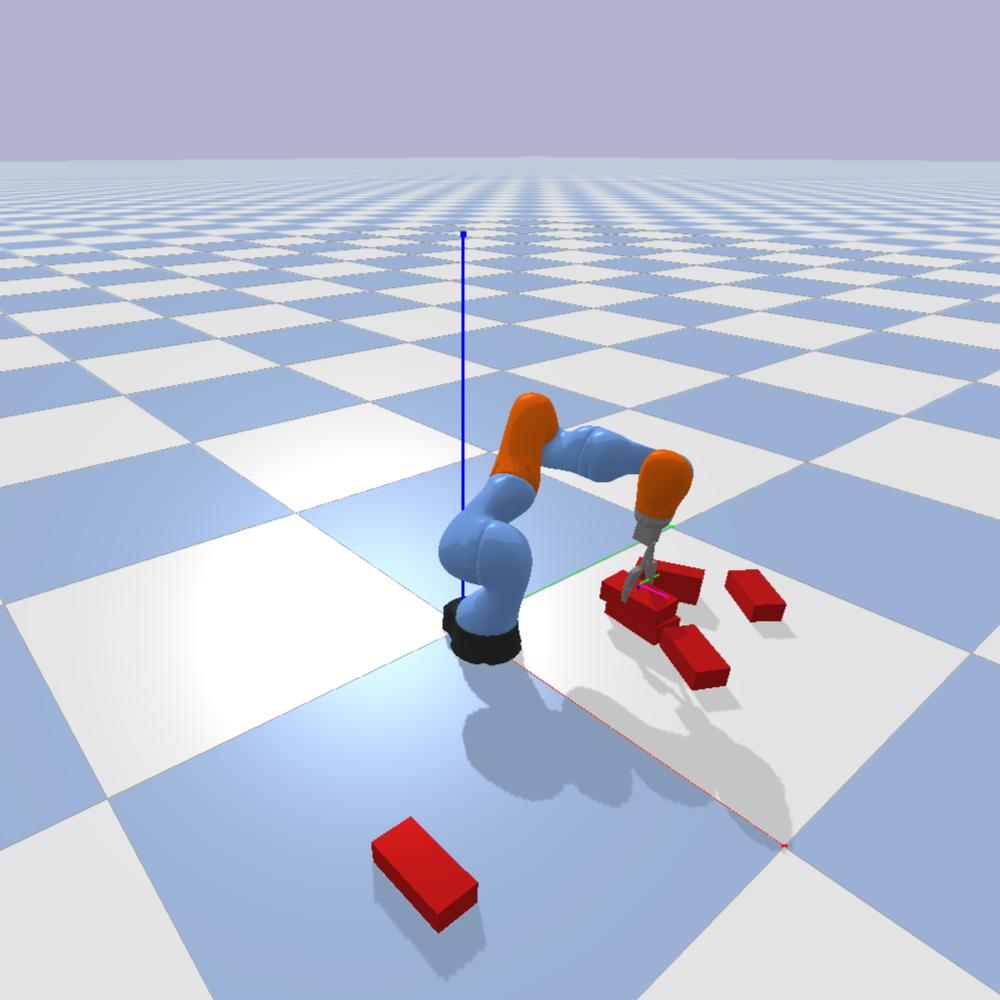}\hfill
  \end{subfigure}

  \vspace{0.5em}

    \begin{subfigure}{0.5\textwidth}\centering
    \textbf{Multi-Agent (Ours)}\\[0.3em]
\includegraphics[width=0.31\linewidth]{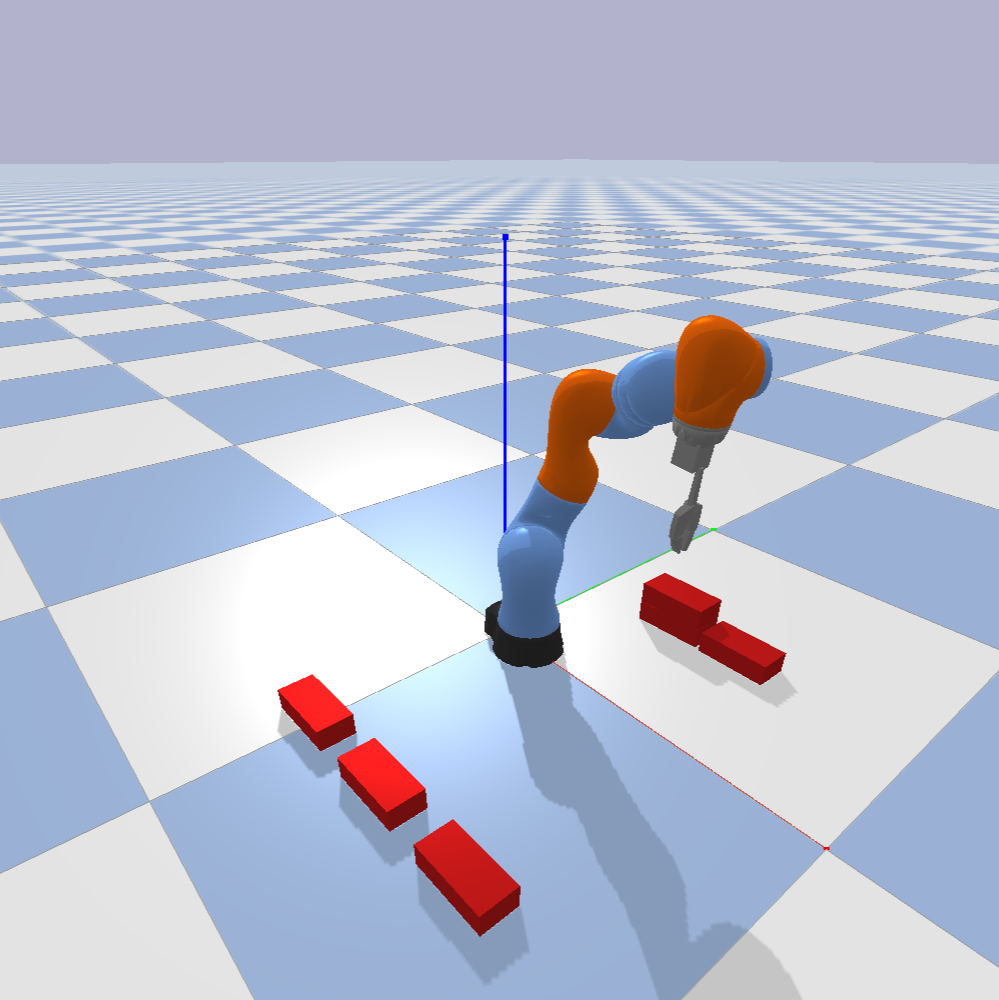}\hfill \includegraphics[width=0.31\linewidth]{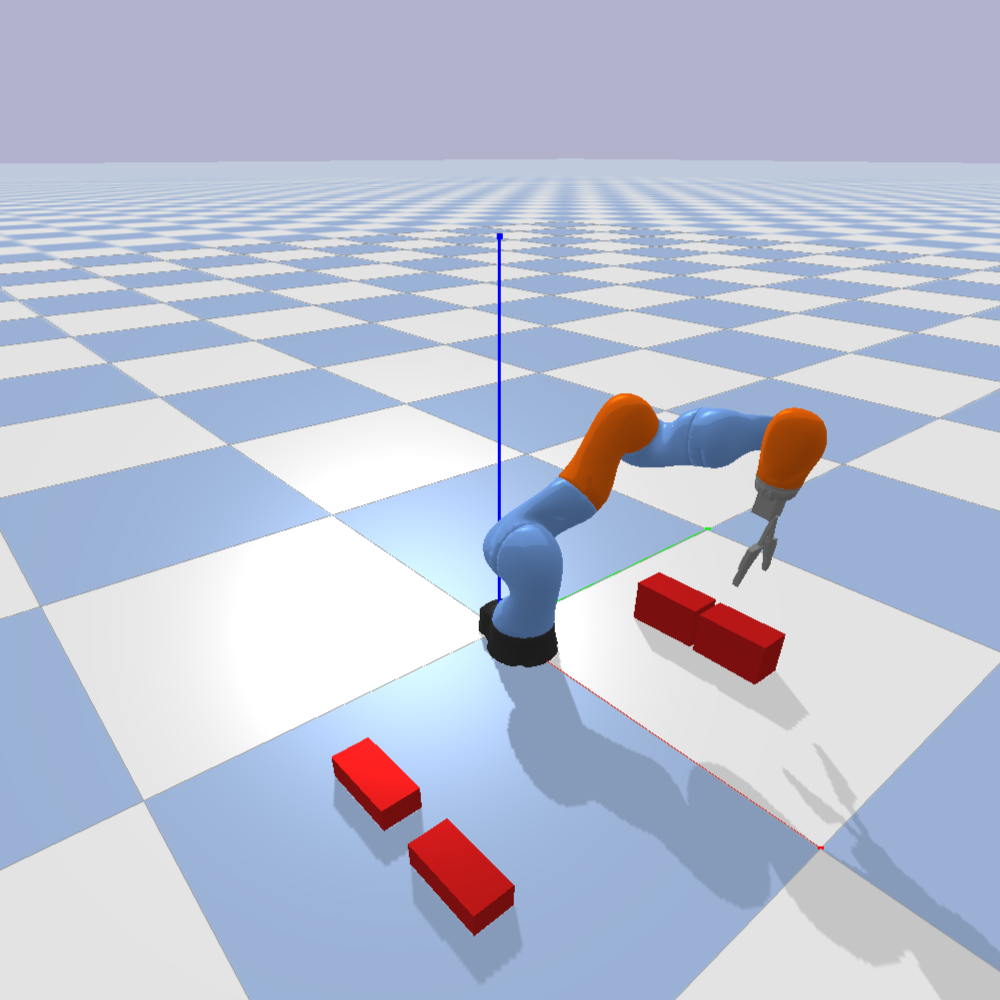}\hfill
\includegraphics[width=0.31\linewidth]{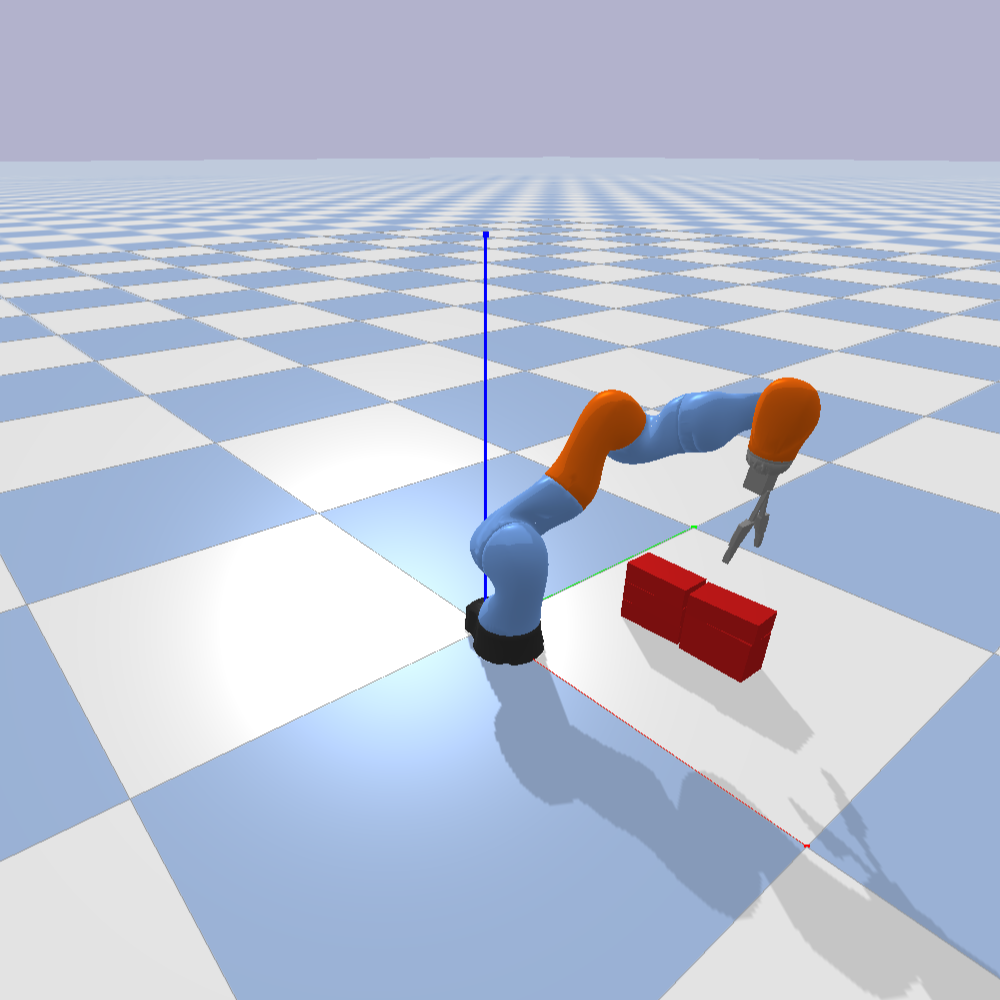}\hfill
  \end{subfigure}

  \caption{Ablation visualizations for single agent. Rows depict single-agent vs multi-agent. Columns show key stages of the placement cycle according to the timestamp from left to right. The figure shows that the single agent toppled the structure and failed to complete the whole wall.}
  \label{fig:ablation-viz}
\end{figure}

\subsection{Baseline and Ablation Study Methods}
\label{subsec:baselines}
All methods receive the same perception inputs and target brick stack pattern. All methods share an identical waypoint interface for low-level actuation. Thus, differences in outcomes stem from the reasoning layer, not from the low-level control stack.
\subsubsection{Controller baseline}\label{controller}
We compare our method, ActionReasoning, with a classical controller baseline that follows a fixed hand-scripted method. The baseline does not use thresholded contact events or environment collision checks. The reason is methodological: in our approach, these capabilities are implemented via LLM-driven tool calls, e.g., contact and collision queries, that do not require substantial domain-specific coding. However, the equivalent event handling in a traditional controller would require extensive bespoke code and expert tuning. To keep the engineering burden comparable across methods, we therefore exclude such event-based modules from the classical baseline and restrict it to a minimal hand-scripted pipeline. Therefore, our comparison reflects a realistic scenario in which an engineer with comparable expertise uses (i) a minimal scripted baseline or (ii) an LLM-driven reasoner that reduces low-level coding while increasing functional robustness through high-level tool calls and prompts.

\subsubsection{Single-Agent ablation} To test whether multi-step agents are necessary, the single agent variant merges the prompts of all six agents into a single LLM call per waypoint, removing stage-wise gating.

\subsection{Experiment Results}
\label{subsec:results}
\subsubsection{Controller baseline}
Table~\ref{tab:baseline-compare} shows that our method markedly improves geometric accuracy over the classical controller. The mean rotation error drops from 1.004 cm to 0.703 cm (an \(30.0\%\) reduction), the mean center-offset drops from 4.314 cm to 0.637 cm (an \(85.2\%\) reduction), and the 3D box overlap increases from 0.3838 to 0.8803 (\(129\%\) increase).
The sequences in Fig.~\ref{fig:baseline-viz} corroborate the metrics: 
the baseline exhibits cumulative lateral drift and premature release during the place phase, 
leading to uneven stacks, whereas our method maintains consistent vertical descent, 
event-triggered release, and uniform brick spacing. Because both methods share the same perception inputs and low-level controller, these gains can be attributed to the proposed multi-agent physical reasoning, including thresholded contact detection and collision-gated releases as mentioned in Section \ref{controller}.

\subsubsection{Single-Agent ablation} We evaluate the single-agent variant under the wall-stacking setting. This ablation consistently underperforms the proposed staged pipeline, supporting the need for explicit role specialization and stage-wise gating. As shown in Fig. \ref{fig:ablation-viz}, the single-agent controller places the first four bricks with noticeably larger placement error and does not robustly handle the final two, frequently colliding with the structure and toppling the wall. We attribute this to the lack of inter-stage verification in the single-agent method, which allows early errors to propagate. This ablation underscores the necessity of multi-stage reasoning.

\section{CONCLUSIONS}
\label{sec:conclusion}

In this paper, we have introduced ActionReasoning, an LLM-driven approach to robotic manipulation that performs physical reasoning in \( \mathrm{SE}(3) \). By conceptualizing robot control as a set of specialized agents with explicit roles and gates, and by exposing LLMs to both an environment state \( S_t \) and callable physics/tool knowledge \( F \), the system can reference relevant formulas and invoke verified functions to generate executable waypoints \( w_{t+1} \in \mathrm{SE}(3) \). In the brick-laying domain, this design bridges the gap between understanding and doing: the LLM reasons over 3D scene geometry, object properties, and task constraints, while the controller handles trajectory interpolation and low-level servoing. Experiments in simulation demonstrate that the proposed multi-agent pipeline yields robust placement behavior, and ablations indicate that staged, role-specific reasoning is essential.

{\bf Future work.} We will extend this framework beyond brick stacking to a broader set of complex unstructured construction site tasks and materials with varied physical properties: mortar deposition and leveling, block/stone/wood handling, fastening and drilling, compliant insertion, and cluttered-scene assembly with dynamic obstacles. Ultimately, our goal is a general-purpose construction robot capable of executing diverse tasks on-site and adapting to different objects, tools, and materials, so that a single robot can complete end-to-end building workflows with minimal low-level coding of specific task.

\bibliographystyle{./IEEEtran} % use IEEEtran.bst style
\bibliography{IEEEexample}

\end{document}